\documentclass{article}

\usepackage[utf8]{inputenc} \usepackage{hyperref}       \usepackage{url}            \usepackage{booktabs}       \usepackage{amsfonts}       \usepackage{nicefrac}       \usepackage{microtype}     \usepackage{natbib}
\usepackage{graphicx}
\usepackage{amssymb,amsmath,amsthm}

\newtheorem{theorem}{Theorem}
\newtheorem{lemma}{Lemma}
\newtheorem{proposition}{Proposition}
\theoremstyle{remark}

\def\ms{m_1}
\def\mt{m_2}
\def\hms{\hat{m}_1}
\def\hmt{\hat{m}_2}
\def\Ss{\Sigma_1}
\def\St{\Sigma_2}
\def\hSs{\hat{\Sigma}_1}
\def\hSt{\hat{\Sigma}_2}
\def\hA{\hat{A}}
\def\ph{^{\frac{1}{2}}}
\def\pmh{^{-\frac{1}{2}}}

\def\E{\mathbb{E}}

\def\Ps{{\mathcal{P}_s}}

\def\Pt{{\mathcal{P}_t}}

\usepackage{xcolor, soul}

\newcommand\rf[1]{{\color{black} #1}}

\newcommand\supp[2]{ #1}

\title{Concentration bounds for linear Monge mapping estimation and optimal transport domain adaptation}

\author{%
  Rémi Flamary\thanks{Both authors contributed equaly.}, 
Karim Lounici$^*$, 
   André Ferrari
}

\begin{document}

\maketitle








\begin{abstract}
  This article investigates the quality of the estimator of the linear Monge mapping 
  between distributions. We provide the first
 concentration result on the linear mapping operator and prove a sample complexity of
 $n^{-1/2}$ when using empirical estimates of first and second order moments. 
  This result is then used to derive a generalization bound for domain adaptation
  with optimal transport. As a consequence, this method
  approaches the performance of theoretical Bayes predictor under mild conditions
  on the covariance structure of the problem.
  We also discuss the
 computational complexity of the linear mapping estimation and show that when
  the source and target  are stationary the mapping is a convolution that can be
 estimated very efficiently using fast Fourier transforms. Numerical
 experiments reproduce the behavior of the proven bounds on simulated and real
 data for mapping estimation and domain adaptation on images. 
\end{abstract}



\begin{abstract}

This article investigates the quality of the estimator of the linear Monge mapping 
 between distributions. We provide the first
concentration result on the linear mapping operator and prove a sample complexity of
$n^{-1/2}$ when using empirical estimates of first and second order moments. 
 This result is then used to derive a generalization bound for domain adaptation
 with optimal transport. As a consequence, this method
 approaches the performance of theoretical Bayes predictor under mild conditions
 on the covariance structure of the problem.
 We also discuss the
computational complexity of the linear mapping estimation and show that when
 the source and target  are stationary the mapping is a convolution that can be
estimated very efficiently using fast Fourier transforms. Numerical
experiments reproduce the behavior of the proven bounds on simulated and real
data for mapping estimation and domain adaptation on images. 
\end{abstract}

\section{Introduction}

Optimal Transport (OT) aims at finding the solution of least effort to move mass from one distribution to another. 
It is a fundamental problem strongly related to physics and has been investigated 
by mathematicians since the introduction of the problem by Monge
\citep{monge1781}. One major result by
\citep{brenier1991polar} proved the existence and derived smoothness properties of optimal transport mapping
between continuous distributions. 

\paragraph{Optimal Transport in statistical learning.}

The interest for OT and the related Wasserstein distance have been growing in
machine learning in recent years \citep{arjovsky17a,courty2016optimal,Frogner15} as OT
provides a convenient tool to compare the geometry of
distributions. In addition, the Wasserstein distance is one of the few divergence that can be
applied (and sub-differentiated) on empirical distribution with no need for kernel smoothing as done in
MMD \citep{gretton2012kernel}. One recent applications of OT concerns the training of Generative Adversarial Networks, a particularly difficult optimization problem where the Wasserstein distance
has been used to provide meaningful gradients \citep{arjovsky17a,liu2018two,lei2019geometric}. OT has also been used in other learning problems
such as unsupervised Domain Adaptation (DA) that aims at training a classifier
that perform well on an unlabeled target dataset using information from a
related but different labeled source dataset as discussed below in more details. 

The rising
interest of the machine learning community has been rendered possible thanks to the   
recent development of efficient optimization techniques. For instance entropic regularization
\citep{CuturiSinkhorn,benamou2015iterative} has lead to new efficient algorithms
that can scale to large datasets and even opened the door to stochastic
optimization \citep{genevay2016stochastic,seguy2017large}. 

In Machine Learning applications, the distributions are typically unknown and we only have access to finite sample 
realizations. This raises the question of estimation of the OT mapping between 
the underlying distributions based only on finite samples. The problem of estimating a continuous mapping has surprisingly received very little attention in the literature. 
\cite{stavropoulou2015parametrization} proposed a two-step procedure for estimating a continuous mapping from discrete distributions. 
\cite{perrot2016mapping} built up upon this approach by adding ridge-type regularization to prevent overfitting with small samples. 

\citep{fournier2015rate,weed2017sharp} derived the rate of convergence of the empirical distribution to its population counterpart with the Wasserstein
distance and showed in particular that these rates suffer from the \textbf{curse of dimensionality}.  \citep{hutter2019minimax} highlighted that the Monge mapping estimation problem belongs to the class of nonparametric statistical problems that are also known to suffer from the
curse of dimensionality. More specifically, they derived a minimax lower bound of the order $O(n^{-1/d})$ where the dimension of the problem $d$ is typically large in modern ML applications. 
Their result guarantees that OT is hopeless without additional structure assumption.
In this paper, we investigate the impact on the estimation rate of adding structure assumption on OT mapping. 
We prove that assuming linear OT structure can help \textbf{both with statistical and computational performances}
as we obtained a practical and easy to compute estimator that attains faster \textbf{dimension free parametric rate of estimation} $O(n^{-1/2})$.

\paragraph{Domain adaptation.}

Domain Adaptation (DA) is a problem in machine learning that is part of the larger Transfer Learning
family. The main problem that is addressed in unsupervised DA is how to predict classes on a new
(target) dataset that is different from the available (source) training dataset. In order to train a classifier
 that works well on the target data, we have access to labeled samples $(X^s_i
, X^s_i )$ drawn from the source joint
feature/target distribution $\mathcal{P}_s$ 
(whose feature marginal is defined as $\mu_s$) and only to feature samples $X^t_j$ from the marginal $\mu_t$ of the joint target distribution $\mathcal{P}_t$.

The literature on domain adaptation is extensive with numerous different approaches.We will discuss briefly the
main approaches before introducing our contribution.
A popular approach is based on re-weighting schemes \citep{sugiyama2008direct} where the main
idea is to re-weight the
samples in order to compensate for the discrepancy between source and target distributions. A classifier can then be estimated by minimizing
the weighted source distribution. These approaches have been shown to work very well in numerous cases but a classic failure scenario would be when the distributions do not overlap.
Another popular DA approach is known as subspace method. When the datasets are
high-dimensional, this approach assumes that there exists a subspace that is discriminant in both source and
target domains. This subspace can then be used to train a robust classifier. A standard approach is
to minimize the divergences between the two projected distributions onto lower-dimensional spaces \citep{Si10}.  Source
labels (when available) can be integrated into this approach to produce a subspace that preserves the class discrimination after projection \citep{long2014transfer}.
Finally the last approach aims at aligning the source and target distribution through a more complex
representation than linear subspace. \cite{Gopalan14} proposes to align the distributions by following
the geodesic between source and target distributions. Geodesic flow kernel \citep{gong12} aims at adapting
distribution using a projection in the Grassmannian manifold and computing kernels using the geodesic flow
in this manifold. Note that all the methods above aim at finding a way to compensate for the change
in distribution between the source and target domains. The same philosophy has been investigated for
neural networks where the feature extraction aims at being indistinguishable between the different domains using
Domain Adversarial Neural Network (DANN) \citep{ganin2016}. Other approaches have aimed at minimizing
the divergence between the distributions in the NN embedded space using covariance matrix alignment
(CORAL) \citep{Sun2016}, minimization of the Maximum Mean Discrepancy (MMD) \citep{adda} or Wasserstein distance
in the embedded space \citep{shen2018wasserstein}. More references can be found in \citep{csurka2017domain} about visual adaptation that has
been very active recently.

\begin{figure}[t]
    \centering
    \includegraphics[width=.99\linewidth]{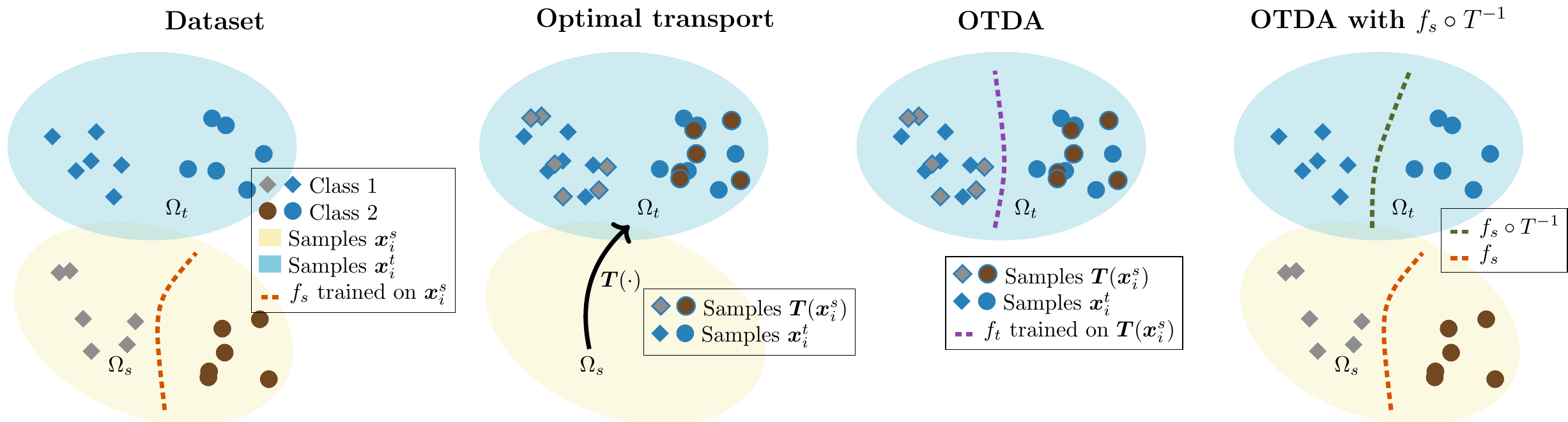}
    \caption{Illustration of Optimal Transport for Domain Adaptation (OTDA) from \citep{courty2016optimal}}
    \label{fig:otda_courty}
\end{figure}

\paragraph{Optimal Transport for Domain Adaptation (OTDA).}
Recently, \cite{courty2016optimal} suggested to use OT for Domain Adaptation. This approach assumes 
that there exists a transformation between the source/target distributions. If this transformation can be accurately estimated, then the knowledge from the source domain can be accurately transferred to the target domain before learning the classifier. While there is in theory several possible mappings between source and target domains, this approach chooses the optimal
transport Monge mapping since it corresponds to the least effort (a sound approach often
found in nature). This approach has been implemented with success in \citep{courty2016optimal,perrot2016mapping,seguy2017large,courty2017joint}.  Note that as illustrated in Figure~\ref{fig:otda_courty}, target classification can be performed in the source or target domains as discussed in Section \ref{sec:dabound}.

OTDA has also been used in several biomedical applications.
\cite{gayraud2017optimal} have applied OTDA to the problem of P300 detection in
Brain Computer Interfaces. It allowed to adapt between different subjects and can
potentially decrease the time needed for calibration.  OTDA was also applied to
the problem of Computer Aided Diagnostic of Prostate cancer from MRI in
\citep{gautheronadaptation} to adapt between patients. Finally
\cite{chambon2018domain} showed that OTDA can improve the performance of sleep
stage classification from ElectroEncephaloGrams.

Despite numerous practical applications, little is known about the statistical performances of OTDA approaches. {
 \cite{courty2016optimal} and
\cite{redko2017theoretical} derived
generalization bounds that include a divergence term between the
source and target distributions close to that of
\cite{ben2010impossibility}. Because of this term, those generalization bounds
require source and target
distributions to be similar in order to achieve domain adaptation.}
In this paper, we derive explicit generalization bounds for OTDA under milder
conditions since OT and its corresponding Monge mapping has the ability to align distributions.
The core of our analysis is a new concentration bound for the estimation of Linear Monge mapping between two distributions based on finite samples, which is of interest in itself.

\paragraph{Contributions.}
\begin{itemize}
\item We prove that adding and exploiting structure assumptions on the OT mapping leads to significant improvements both in terms of statistical bounds and computational complexity. We obtained a dimension-free parametric rate  $O(n^{-1/2})$ for the estimation for the OT mapping  with a polynomial computational complexity in the dimension $d$ of the problem.

\item We derive new generalization bounds for the OTDA problem. Interestingly, these bounds no longer depend on the divergence between the source and target distributions if we assume that there exists a transport map from the source and target distributions.

\item We performed numerical experiments on simulated and real datasets that confirm the interest of our approach.
\end{itemize}

\paragraph{Organization of the paper.}

The rest of the paper is organized as follows. In Section~\ref{sec:main}, we focus on the estimation of a linear Monge mapping arising from transport between Gaussian distributions \citep{takatsu2011wasserstein}. We prove that the closed form solution of the linear Monge mapping is also valid for any Borel 
distributions with finite second order moments. Next, we derive in Theorem \ref{thm:ot_Monge} our main concentration bound for linear Monge mapping based on
finite samples of sub-gaussian distributions. This result is then used in Section \ref{sec:dabound} to derive a new generalization bound
for OT Domain Adaptation. In Section \ref{sec:numerical_exp},
we carry out numerical experiments
for the estimation of the
monge mapping and the performance of OTDA.

\paragraph{Definitions and notations.}
In what follows, for any symmetric positive definite matrix $B$, we denote by
$\lambda_{\min}(B)$, $\lambda_{\max}(B)$ the minimum and maximum eigenvalues of
$B$ respectively. We also define the effective rank of $B$ by $\mathbf{r}(B) =
\frac{\mathrm{tr}(B)}{\lambda_{\max}(B)}$ where $\mathrm{tr}(B)$ is the trace of
$B$. By abuse of notation, $\|\cdot\|$ refers either to the $l_2$-norm of a
vector or to the operator norm of a matrix.  We also define the condition number
of $B$ as $\kappa(B) = \frac{\lambda_{\max}(B)}{\lambda_{\min}(B)}$. 
Finally we use the binary operators $\vee$ and $\wedge$
to denote maximum and minimum respectively.

\section{Linear Monge mapping estimation and concentration}
\label{sec:main}

\paragraph{Linear Monge mapping between Gaussian distributions} Let $\mu_1=\mathcal{N}(m_1,\Sigma_1)$ and $\mu_2=\mathcal{N}(m_2,\Sigma_2)$ be
two distributions on $\mathbb{R}^d$. In the remaining we suppose that both $\Sigma_1$ and
$\Sigma_2$ are symmetric positive definite.
The Monge mapping for a quadratic loss between $\mu_1$ and $\mu_2$ can be expressed as 
\begin{equation}
    T(x)=m_2+A(x-m_1)\label{eq:map}
\end{equation}
with 
\begin{equation}
    A=\Sigma_1\pmh\left(\Sigma_1\ph\Sigma_2\Sigma_1\ph\right)\ph\Sigma_1\pmh=
    A^T\label{eq:Amap}
\end{equation}
This is a well known fact in the Optimal Transport literature \citep{givens1984class,mccann1997convexity,takatsu2011wasserstein,bhatia2018bures,malago2018wasserstein}. See also \citep[Remark 2.31]{peyre2019computational}.
Note that the matrix $A$ is
actually the matrix geometric mean between $\Sigma_1^{-1}$ and $\Sigma_2$: $A = \Sigma_1^{-1}\# \Sigma_2$.

\paragraph{Linear Monge mapping between general distributions} 

The linear mapping between Gaussian distribution discussed above is very elegant
but real life data in machine learning seldom follow Gaussian distribution
(especially classification problems that are at best a mixture of Gaussian). It turns out that  the optimal linear transport between arbitrary distributions with finite second order moment is the same as in the Gaussian case. This result was essentially proved in \citep{Dowson1982} where a lower bound on the Fr\'echet distance between distributions is established. The only missing ingredient was the seminal result on optimal transport that was proved by \citep{brenier1991polar} several years later. Combining these two results, we obtain Lemma \ref{lem:ot} below.

\begin{lemma}\label{lem:ot} Let $\mu_1$ and $\mu_2$ be two Borel probability measures with
finite second order moments with
expectations $m_1,m_2$ and positive-definite covariance operators $\Sigma_1$,
$\Sigma_2$ respectively and such that $\mu_2=\tilde T_\#\mu_1$ for an affine
$\tilde T(x)=Bx+c$ with $B$ symmetric positive definite. Then the optimal
transport mapping is $\tilde T=T$ where $T$ is defined by (\ref{eq:map})-(\ref{eq:Amap}).     \label{thm:subgauss}
\end{lemma}

\begin{figure}[t]
    \centering
    \includegraphics[width=.99\linewidth]{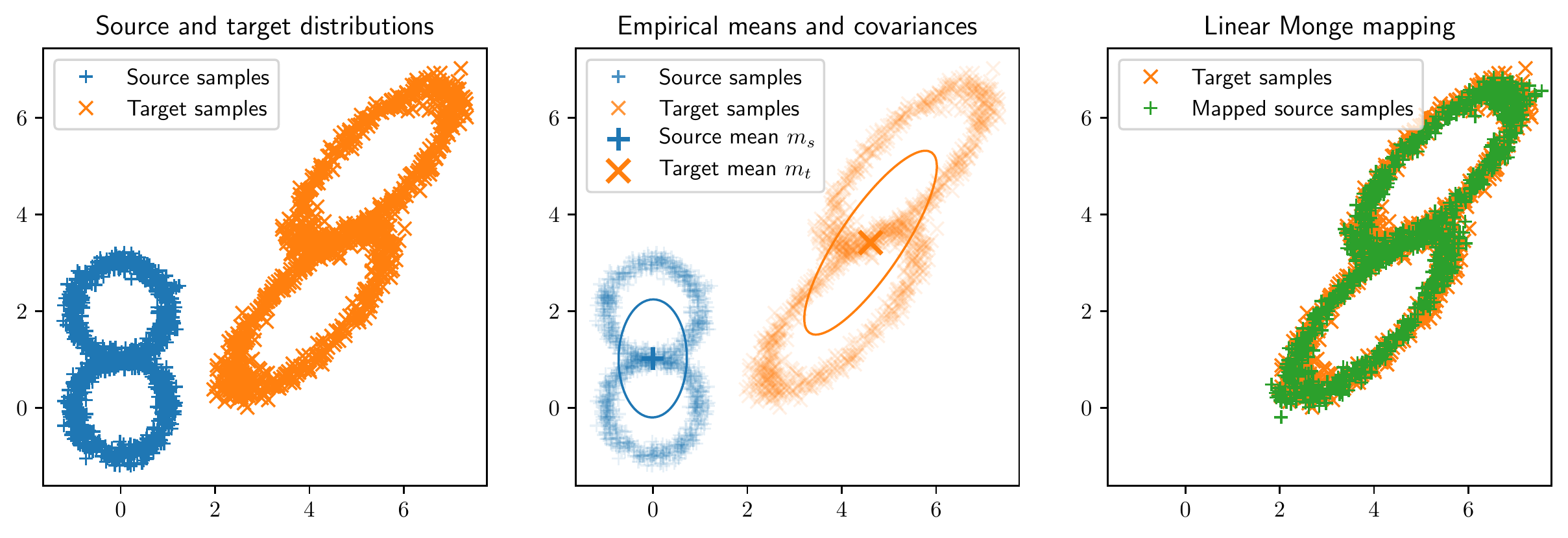}\vspace{-3mm}
    \caption{Example of linear Monge mapping estimation between empirical distributions. (left) 2D source and target distributions. (middle) Estimations for means and covariances of the distributions. (right) resulting linear mapping where green samples are the mapped source samples.}
    \label{fig:example_map}
\end{figure}

\paragraph{Estimation bounds for the Monge mapping} In practice the
distributions $\mu_1$, $\mu_2$ are unknown and we have only access to independent samples $\mathbb{X}_1 = \left\lbrace X_1^s,\ldots,X_{n_1}^s \right\rbrace$ and $\mathbb{X}_2 = \left\lbrace X_1^t,\ldots,X_{n_2}^t \right\rbrace$
where the $X_i^s$ are i.i.d. with distribution $\mu_1$ and the $X_j^t$ are i.i.d. with distribution $\mu_2$. In this case, the linear Monge
mapping can be estimated using
empirical means and covariances $\hms,\hmt,\hSs,\hSt$ based on $n_1$ and
$n_2$ samples respectively. 
Hence we define the empirical linear Monge mapping as
\begin{equation}
\hat T{(x)} = \hat T _{(\mathbb{X}_s,\mathbb{X}_t)}{(x)} = \hmt + \hat A (x-\hms),
\label{eq:That}
\end{equation}
where $\hat A$ comes from (\ref{eq:Amap}) where the covariances are replaced
by their empirical counterpart.  Note that this requires $\hat\Sigma_1$ to be nonsingular, in other words we need $n_1 \geq d$ observations in the source domain to compute this estimator. An illustration of this method for a highly
non-Gaussian distribution can be seen in Figure \ref{fig:example_map}. {We can
clearly see here that under the assumptions in Lemma \ref{thm:subgauss},
we can recover the Monge mapping and align very well complex distributions.}

Let $T$ and $T'$ be two mappings, we define the $L_2$-divergence between mapped distributions $T_\#\mu_1$ and $T'_\#\mu_1$ as
\begin{align}
    d(T,T')  =  \mathbb{E}_{x\sim \mu_1} \left[ \| T(x) - T'(x) \|  \right].
    \label{eq:distmapping}
\end{align}

In the next theorem, we prove a bound for the error of estimation of $T$ by $\hat T$.
\begin{theorem}\label{thm:ot_Monge}
  Let $\mu_1$ and $\mu_2$ be sub-Gaussian distributions on $\mathbb{R}^d$ with expectations $m_1,m_2$ and positive-definite covariance operators $\Sigma_1$, $\Sigma_2$ respectively. We assume furthermore that 
  \begin{align}\label{cond-1-Sigma_j}
  c<\min_{j=1,2}\lbrace( \lambda_{\min}(\Sigma_j)\rbrace  \leq  \max_{j=1,2} \lbrace \lambda_{\max}(\Sigma_j)\rbrace \leq C,
  \end{align}
          for some fixed absolute constants $0<c\leq C<\infty$. We also assume that
  \begin{align}\label{cond-2-Sigma_j}
 n_1 \geq d,\quad  n_2 \geq C' \mathbf{r}(\Sigma_2),
  \end{align}
  for some sufficiently large numerical constant $C'>0$.
  
  Then, for any $t>0$, we have with probability at least $1-e^{-t}-\frac{1}{n_1}$,
  \begin{align}
  d(T,\hat{T}) \leq  C''   \left(      \sqrt{\frac{\mathbf{r}(\Sigma_1)}{n_1}} \vee \sqrt{\frac{\mathbf{r}(\Sigma_2)}{n_2}} \vee   \sqrt{\frac{t}{n_1 \wedge n_2}} \vee \frac{t}{n_1\wedge n_2} \right) \sqrt{\mathbf{r}(\Sigma_1)},
  \end{align}\label{thm:err_map}
  where $C''>0$ is a constant independent of $n_1,n_2,\mathbf{r}(\Sigma_1),\mathbf{r}(\Sigma_2),d$.
  \end{theorem}
The detailed proof is provided in Appendix \ref{sec:proof}. 

 \subsection*{\textbf{Comments}}
\label{sec:}

\begin{itemize}

\item \textbf{Statistical performance.} This result is one of the first bound on the quality of an estimated continuous Monge mapping from empirical distribution. By exploiting the additional linear structure of the Monge mapping, we obtain in (\ref{thm:err_map}) a faster dimension-free estimation rate of the order $O(n^{-1/2})$ when $n=n_1=n_2$ which compares  favorably to the $O(n^{-\frac{1}{d}})$ obtained with the more general but not
  computationally feasible estimator in \cite{hutter2019minimax}.
  Note that this result 
  also provides a convergence rate for the generalization bound in \cite[Eq. (13)]{perrot2016mapping} in the linear case where the term $d(T,\hat{T})$
  appeared in the bound but was not studied.

\item \textbf{Computational complexity: general covariance matrices.}  The mapping is estimated from empirical distributions by
using the empirical version of the means and covariances in
\eqref{eq:map}-\eqref{eq:Amap}. The complexity of estimating those parameters in
$O((n_1+n_2)d^2)$ which is linear \emph{w.r.t.} the number of samples but quadratic
in dimensionality $d$ of the data. Eq. \eqref{eq:Amap} also requires the
computation of matrix square root and inverse which are $O(d^3)$ leading to a final complexity
of $O((n_1+n_2)d^2+d^3)$. This complexity scales well with the number of
training samples but not with the dimensionality of the space. 

\item \textbf{Computational complexity: convolutional Monge mapping on signals and images.} 
When the data samples are temporally or spatially stationary signal or images,
it is common practice to approximate  their Toeplitz or
block-Toeplitz covariance matrices by circulant matrices (for large values of dimension $d$). Indeed, a remarkable property of circulant matrices is that their spectral decomposition can be computed with a Discrete Fourier Transform (DFT):
$\Sigma_1 = F D_1 F^\ast$, $\Sigma_2 = F D_2 F^\ast$, 
\cite{Gray2005}.  This is of particular interest because the linear
operator in \eqref{eq:Amap} whose computational complexity in the general case is $O(d^3)$ becomes a convolution operator with frequency response
$D = D_2^{\frac{1}{2}}D_1^{-\frac{1}{2}} $: 
\begin{equation}
A = FDF^\ast \label{eq:mapconv}
\end{equation}
that can be computed efficiently in the Fourier domain using the Fast Fourier
transform (FFT) algorithm. Thus this additional assumption leads to  a significant speedup of the estimation of the mapping on large dataset and large samples (images
or signals).
The speedup of the FFT leads to a final computational cost
of $O((n_1+n_2)d\log(d))$ {to estimate $D$} that is greatly reduced compared to the general linear
case discussed above. Note that in this case
in order to use the FFT we suppose that the linear mapping
operator is a positive definite circular convolution operator which can
introduce artifacts at the border of images. 

\item \textbf{Regularization.} We suppose in all our theoretical results
that the covariance matrices are positive definite. In practice this assumption can be false for instance when the data lies in a subspace. A
standard practice is to replace the empirical covariance matrices $\hat\Sigma_j$ by
$\tilde \Sigma_j=(1-\alpha)\hat\Sigma_j+\alpha I$ where $\alpha\geq 0$ and  $I$ is the identity
matrix. In our numerical experiments we did not use this regularization in the
simulated examples but needed to use it with $\alpha=10^{-6}$ on the real life image data.
\end{itemize}

\section{Domain adaptation generalization bound}
\label{sec:dabound}

Now we focus on the problem of domain adaptation where we have access to data
from a source joint feature/label distribution $\Ps$ but want to predict well on
a target joint distribution $\Pt$ where only features with
marginal distribution
$\mu_t$ are available. We consider two pairs of samples $(X^s,Y^s)$ and $(X^t,Y^t)$ with values respectively in $\mathcal{X}^s \times \mathcal{Y}$ and $\mathcal{X}^t \times \mathcal{Y}$ with respective joint distributions $\Ps$ and $\Pt$. Note that $\mathcal{X}^s$ and $\mathcal{X}^t$ are possibly different as it is typically the case in several DA applications.

We define the risk of a prediction rule $f$ in the source domain as
\begin{equation}
    R_s(f) : = \mathbb{E}_{(X,Y)\sim \Ps}\left[ L(Y,f(X))   \right].
    \label{eq:risk_source}
\end{equation}
where $(X,Y)\in \mathcal{X}^s\times \mathcal{Y}$ and $L\,:\, \mathcal Y \times \mathcal Y \rightarrow \mathbb{R}^+$ is a Lipschitz loss function  \emph{w.r.t.} its second variable. We denote by $M_L$ the Lipschitz constant. For instance we have $M_L=1$ for the hinge loss. The risk on the target domain $R_t(f)$ is defined similarly with expectation \emph{w.r.t.} $\Pt$.
The optimal prediction rule in the source domain is defined as
\begin{align}\label{eq:optf**}
f_*^s := \mathrm{argmin}_{f\,:\,\mathbb{R}^d\rightarrow \mathbb{R}}\quad  R_s(f),
\end{align}
where the minimization is taken over all measurable functions. We assume here
for simplicity that the minimum is attained. Similarly, the optimal prediction
rule on target is defined as $f_*^t$.
Note that although the Bayes prediction rule in the previous display does not depends on the marginal distribution of $X^s$, it is defined on domain $\mathcal{X}^s$. Therefore, when $\mathcal{X}^s$ and $\mathcal{X}^t$ are different, it is not possible to simply estimate a predictor $\hat{f}^s$ on the source domain $\mathcal{X}^s$ and apply it directly to unlabeled points in the target domain. \cite{courty2016optimal} suggested an original approach based on optimal transport to resolve this difficulty.

\paragraph{Optimal Transport Domain Adaptation.} 
Following \cite{courty2016optimal}, we assume that there exists a mapping $m$ between the source and target such that $\Pt=m_\#\Ps$ and that the pusfhorward $m$ can be expressed as $m(x,y)=(T(x),y)$. In other words the
samples in the feature spaces have been transformed by $T$ but have conserved
their label through this transformation. 
This assumption corresponds to a number
of real life situations such as a change in the
acquisition conditions, sensor drifts, thermal noise in signal processing. This
implies that for functions $f$ and $g$ in the source and target domains
respectively :
\begin{equation}
    R_s(f)=R_t(f\circ T^{-1}) \quad \text{ and } \quad R_t(g)= R_s(g\circ T)
    \label{eq:assumptionotda}
\end{equation}
where $T$ is assumed to be invertible.
{Note that \eqref{eq:assumptionotda} and \eqref{eq:optf**} imply that  the best performance in the
source and target domains are equal $ R_s(f_*^s)=R_t(f_*^t)$.}
 This motivated the main idea in \cite{courty2016optimal} that if one can
estimate the mapping $T$, then it
is possible to map the labeled source samples in the target domain with $\hat T$ and train a
classifier $\hat g$ in the target domain using the labels from the original source
samples. This classifier can predict the
labels for new data in the target domain. In the following we investigate the generalization performance of a
similar procedure where we train a classifier $\hat f$ in the source domain and use
$\hat f\circ \hat T^{-1}$ to predict in the target domain (see Fig. \ref{fig:otda_courty}).

We say that the prediction rule $f$ on source domain $\mathcal{X}^s$ is $M_f$-Lipschitz if
$
|f(x) - f(y)| \leq M_f \| x-y  \|,
$
for any $x,y \in \mathcal{X}^s$. We can now state a preliminary result.

\begin{proposition}\label{thm:dapred}
    Let $f$ be a  $M_f$-Lipschitz prediction rule in the source domain. Recall that the loss function $L$ is $M_L$-Lipschitz. Under the OTDA assumption (\ref{eq:assumptionotda}), we have
    \begin{equation}
                 R_t(f\circ \hat T^{-1})\leq R_s(f) + M_f M_L \mathbb{E}_{(X,Y)\sim \mathcal{P}_s} \left[   \| \hat T^{-1}(T(X)) -  \hat T^{-1}(\hat T(Y)) \|   \right]\label{DA-oracleinequality1}
           \end{equation}
    If $\hat T$ is the linear mapping as defined in (\ref{eq:That}) then we have
    \begin{equation}
        R_t(f\circ \hat T^{-1})\leq R_s(f) + M_f M_L  \|\widehat{A}^{-1}\|\, d(T,\hat{T}).
        \label{DA-oracleinequalityMP1}
    \end{equation}
\end{proposition}

 This result means that our transferred rule $f\circ \hat T^{-1}$ will
 perform almost as well in the target domain as the initial rule $f$ in the source
 domain up to a remainder term that depends on the estimation error of the transport mapping. Note that (\ref{DA-oracleinequality1}) is valid for arbitrary transport while (\ref{DA-oracleinequalityMP1}) is specific to linear Monge mapping.

\paragraph{Generalization bound for finite samples.} Note that our goal is to learn a good prediction rule in the target domain $\hat f^t\,:\, \mathcal{X}^t
\rightarrow \mathcal{Y}$ from finite samples. To this end we have access to
respectively $n_1$ and $n_2$
unlabeled samples from the source and target
domains denoted $\mathbb{X}_s$ and $\mathbb{X}_t$ respectively. They will be used to estimate the mapping $\hat T$. We also have access
to $n_{l}$ i.i.d. labeled samples $(X_i^l,Y_i^l)$, $i=1,\ldots,n_{l}$ in the source
domain $\mathcal{X}^s\times \mathcal{Y}$ independent from $\mathbb{X}_s$. We illustrate now how to exploit Theorem \ref{thm:ot_Monge} in the DA prediction problem.

We consider the classification framework in \citep{mendelson}. Let $\mathcal{H}_K$ be a reproducing kernel Hilbert space (RKHS) associated with a symmetric nonnegatively definite kernel $K\,:\, \mathbb{R}^d \times \mathbb{R}^d \rightarrow \mathbb{R}$ such that
for any $x\in \mathbb{R}^d$, $K_x(\cdot) = K(\cdot,x) \in \mathcal{H}_K$ and
$f(x) = \langle f(x),K_x \rangle_{\mathcal{H}_K}$ {for all $f \in
\mathcal{H}_K$}. Assume that $f_* \in \mathcal H_K$ and $\|f_*\|_{\mathcal H_K}\leq 1$. We consider the following empirical risk minimization estimator:
\begin{equation}
    \hat f_{n_{l}} := \mathrm{argmin}_{\|f\|_{\mathcal{H}_K}\leq 1} \frac{1}{n_{l}}\sum_{i=1}^{n_{l}} l(Y^l_i, f(X^l_i)).
\end{equation}
where we assume that the eigenvalues $(\lambda_k)_{k\geq 1}$ of the integral operator $T_K$ of
$\mathcal{H}_K$ satisfy $\lambda_k \asymp k^{-2\beta}$ for some
$\beta>1/2$ . Let consider for instance Gaussian kernel $K(x,x') = \exp\left(-\frac{\|x-x'\|^2}{\sigma^2}\right)$ with $\sigma>0$. Then we have, for any $f\in \mathcal{H}_K$, 
$
|f(x)  - f(x')| \leq \|f\|_{\mathcal{H}_K}\, d(x,x'),
$
with $d^2(x,x') = K(x,x)+K(x',x')-2 K(x,x') = 2 - 2 \exp\left( - \frac{\|x-x'\|^2}{\sigma^2} \right) \leq 2 \frac{\|x-x'\|^2}{\sigma^2} $. Therefore, this framework guarantees that the predictor $\hat f_{n_{l}} $  is $\sqrt{2}/\sigma$-Lipschitz continuous. We define the excess risk of a predictor $g$ in the target domain as
$$
\mathcal{E}_t(g) = R_t(g) - R_t(f_*^t),
$$
where we recall that $f_*^t$ is the Bayes predictor in the target domain. We can now state our main result on OTDA.

\begin{theorem}
Let the assumptions of Theorem \ref{thm:ot_Monge} be satisfied. Assume in addition that $ R_s(f_*^s)=R_t(f_*^t)$. Let $\hat T$ be the linear mapping as defined in (\ref{eq:That}). Then we get with probability at least $1-e^{-t}-\frac{1}{n_1}$,
\begin{align}
\mathcal{E}_t(\hat f_{n_l} \circ \hat T^{-1}) &\lesssim   n_l^{-2\beta/(1+2\beta)}  + \frac{t}{n_l} \label{eq:otda_bound}\\
&\small\hspace{.2cm} +  M_L  \left(      \sqrt{\frac{\mathbf{r}(\Sigma_2)}{n_2}} \vee \sqrt{\frac{\mathbf{r}(\Sigma_1)}{n_1}} \vee   \sqrt{\frac{t}{n_1 \wedge n_2}} \vee \frac{t}{n_1\wedge n_2} \right) \sqrt{\mathbf{r}(\Sigma_1)} .\nonumber
\end{align}
\end{theorem}

The above bound proves that
under the mapping assumption, the generalization error of $\hat f_{n_l} \circ \hat
T^{-1}$ converges to the Bayes risk $R_t(f_*^t)$ in the target domain even though we do not have
access to any labeled samples in the target domain. This is to the best of our
knowledge the first theoretical result that leads to such performances on
unsupervised domain adaptation problem. We can get away from the
impossibility theorem of domain adaptation of \citep{ben2010impossibility} thanks
to the strong assumption on the existence of a linear Monge mapping that allows
bypassing the discrepancy between the source and target distributions. Note also that
while this result concerns linear Monge mapping, the proof argument can be easily extended to any mapping estimation such as Virtual Regressive training \citep{perrot2015regressive} that also have a $o(n^{-1/2})$ convergence rate where $n$ is the number of one-to-one mapping samples between source and target domain. Finally the Lipschitzness of the predictor in the classification problem is satisfied for instance if there is no sample in  the  vicinity  of  the  decision  boundary between classes (margin condition). This Lipschitzness condition can be relaxed to cover more cases especially in the classification problem. We could use for instance the notion of probabilistic Lipschitzness introduced in \citep{Urner2011AccessTU}, that is somehow related to the low-noise condition of \citep{mammen1999}.

\section{Numerical experiments}
\label{sec:numerical_exp}

In this section we perform numerical experiments to illustrate our theoretical results. The linear mapping estimation
from \eqref{eq:That} is implemented using  the LinearTransport class from the Python
Optimal Transport library \citep{flamary2017pot}. The convolutional
mapping has been implemented with FFT as discussed in section \ref{sec:main}.
  
Note that all the code of the numerical experiments will be made freely available upon publication.

\subsection{Convergence of the mapping error and domain adaptation generalization.}
\label{sec:conv_speed}

\paragraph{Linear mapping error between Gaussian distributions.}

In these first numerical experiments, we illustrate the convergence speed in
term of mapping quality as a function of the number of samples in source and
target domains $n=n_1=n_2$. 
To this end, for each dimensionality $d\in\{2,10,100\}$ we first select the
source and target Gaussian distributions by randomly drawing their first and
second order moments. For each experiment we randomly draw for $j\in\{1,2\}$
the means as
$m_j\sim \mathcal{N}(\mathbf{0},10I_d)$ and the true covariance matrices as
$\Sigma_j\sim \mathcal{W}_d(I,d)$ where $\mathcal{W}_d$ denotes Wishart
distributions of order $d$. Once we have selected the population parameters $m_i,\Sigma_i$, we generate $n=n_1=n_2$ i.i.d. samples for each distribution
$\mu_j=\mathcal{N}(m_j,\Sigma_j)$. The empirical means and covariances
are estimated from those samples and then used to estimate the Monge Mapping $\hat
T(x)$ in \eqref{eq:That}.

\begin{figure}[t]
    \centering
    \includegraphics[width=.5\linewidth]{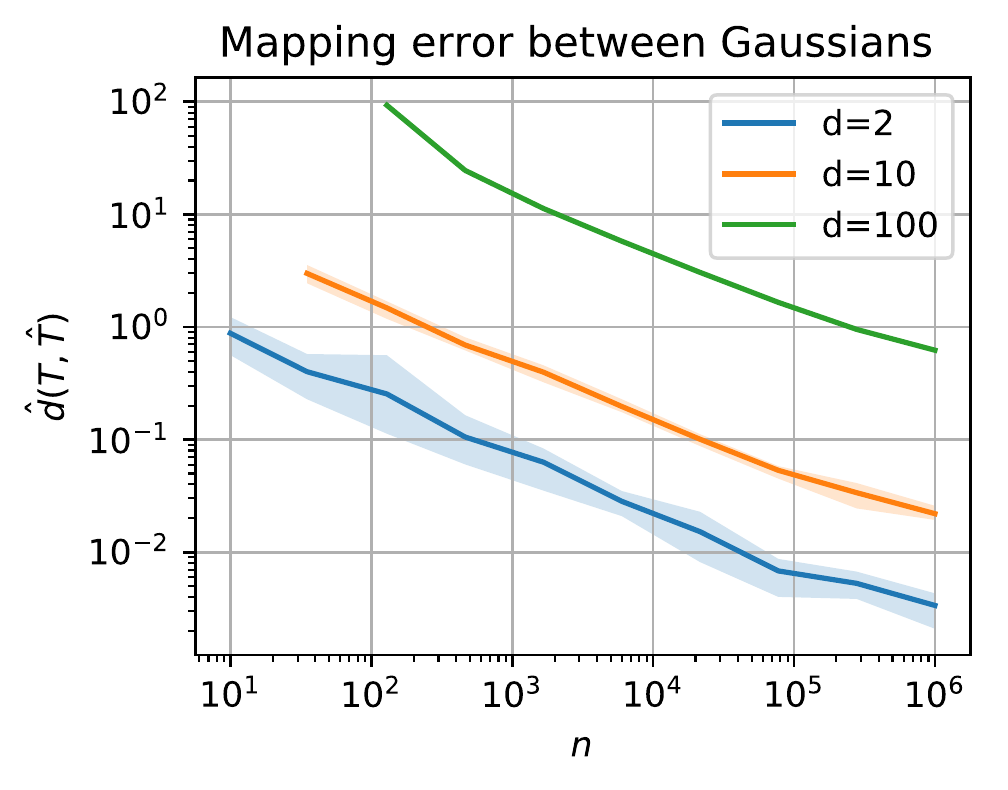}\hspace{-.5mm}
       \caption{Experiments with Gaussian data. Mapping estimation error as a function of $n=n_1=n_2$ for different values of $d$. 
   Colored area corresponds to 10th and 90th percentile.}
   \label{fig:err_map}
\end{figure}

All experiments are repeated $10$ times (Monte Carlo) and the mean mapping error
between the true mapping $T$ and the estimated mapping $\hat T$
is computed on $10^6$ source samples.
Figure \ref{fig:err_map} shows the convergence as a function of $n$ for different
values of  dimensionality $d$. {This log/log plot clearly shows} the slope of
$-\frac{1}{2}$ corresponding to the $O(n^{-\frac{1}{2}})$ convergence speed in
Theorem \ref{thm:err_map}.

\paragraph{Domain adaptation on simulated examples.}
\label{sec:simulated_data}

In the next experiment we reproduce the domain adaptation bounds obtained in section
\ref{sec:dabound}. 
To this end we design a source classification problem where
samples from class + are drawn from $\mathcal{N}(\mathbf{0},\Sigma_0)$ and samples from class -
from $\mathcal{N}(\mathbf{1},\Sigma_0)$ where $\Sigma_0\sim \mathcal{W}_d(I,d)$
is drawn and fixed at the beginning of the experiment. The Bayes classifier in this example is known to have a linear separation.
Therefore we use a Linear
Discriminant Analysis (LDA) classifier to solve this problem. Note that as
discussed in section \ref{sec:dabound} the LDA classifier satisfies the Lipschitz condition since the two class samples are well separated in this experiment. 
In order to design a domain adaptation
problem we generated samples following the source class distributions above and apply a
linear mapping $T(x)=Bx+c$ where $B\sim \mathcal{W}_d(I,d)$ is also computed
and fixed at the beginning and $c=[10,\ldots,10,0,\ldots,0]$ with half of its values
set to 10 and 0 elsewhere. The DA problem above is difficult in the sense that
training on source samples with no adaptation leads to bad performances ($50\%$
accuracy). It also corresponds to a modeling of real life observation where $c$
can be seen as a sensor drift (here half of them will be uncalibrated between
source and target) and the Linear operator $B$ can model a change in the sensor
position.

\begin{figure}[t]
    \centering
       \includegraphics[width=.44\linewidth]{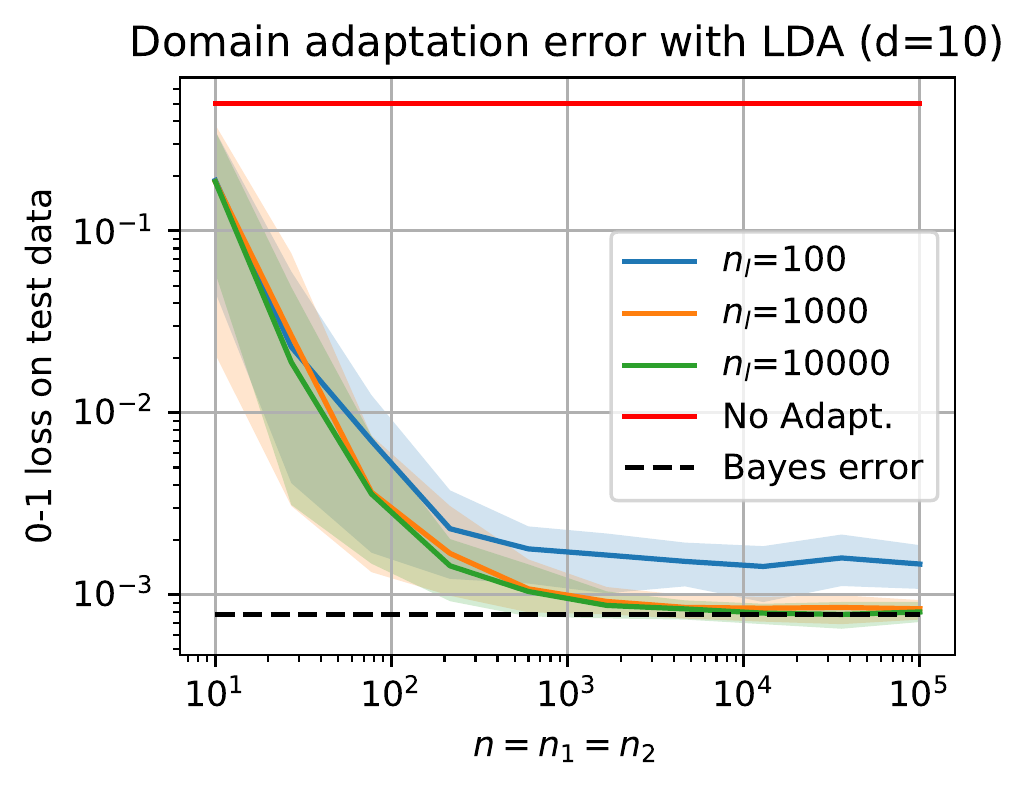}    \includegraphics[width=.44\linewidth]{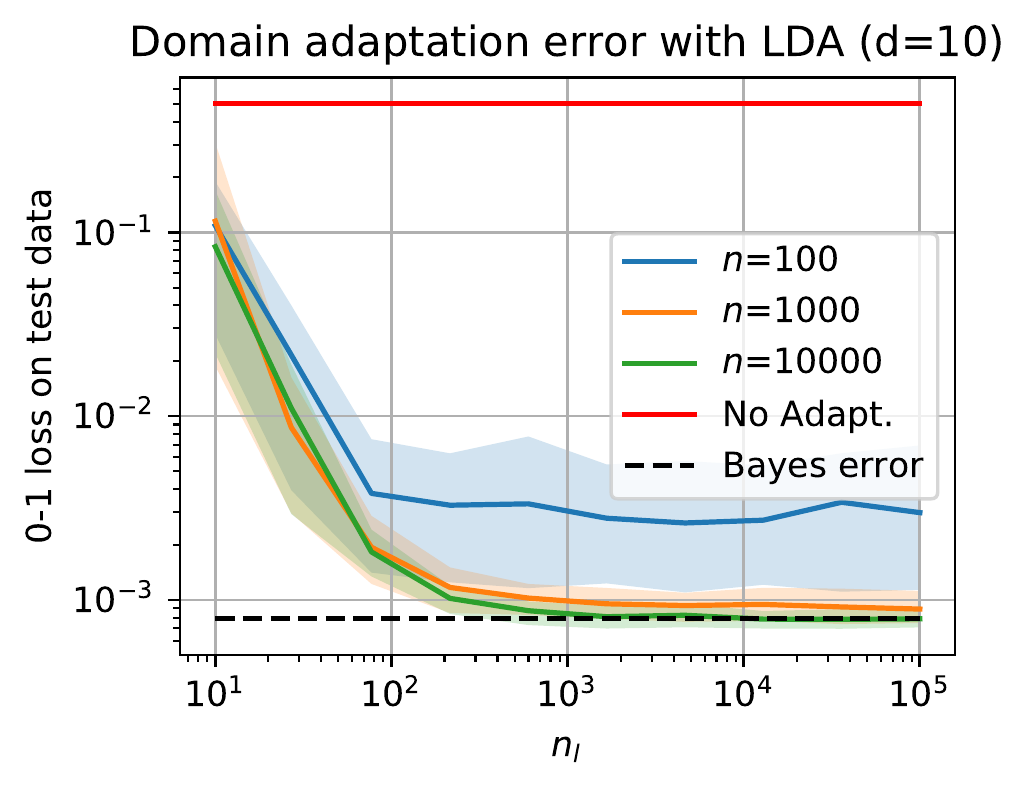}
   \caption{Domain Adaptation experiments with Gaussian data.  (left)  Domain adaptation test error rate
   with Gaussian data as a
   function of $n$ (right) domain adaptation test error rate as a function of
   $n_{l}$.
   Colored area corresponds to 10th and 90th percentile. We also report the
   performance of the Bayes classifier on target and the performance if trained
   on source with no adaptation.}
   \label{fig:err_da}
\end{figure}

All
numerical experiments were performed with $d=10$. We generate $n_{l}$ labeled samples in
the source domain to train the source classifier $\hat f_{n_l}$. We also
generate $n=n_1=n_2$ independent unlabeled samples in both source and target to
estimate the mapping $\hat T^{-1}$. The DA classifier as suggested by 
(\ref{eq:otda_bound}) is the function $\hat f_{n_l} \circ\hat T^{-1} $ and its
accuracy evaluated on $10^6$ independent target samples. 
The average classification error rate on $50$ Monte Carlo realizations is
reported in Figure
\ref{fig:err_da} for different values of $n$ and
$n_{l}$ respectively. We can see in
both plots that when both $n$ and $n_{l}$ become large, the error rate converges to the
Bayes error rate $R_t(f_*^t)$.

\subsection{Convolutional Monge mapping between images}
\label{sec:}

In this section we investigate the estimation of Monge mapping between images
when the mapping is a convolution. To this end we use the well
known MNIST dataset \citep{lecun1998mnist} for both 2D filter estimation and
convolutional domain adaptation.

\begin{figure}[t]
    \centering
    \includegraphics[width=1\linewidth]{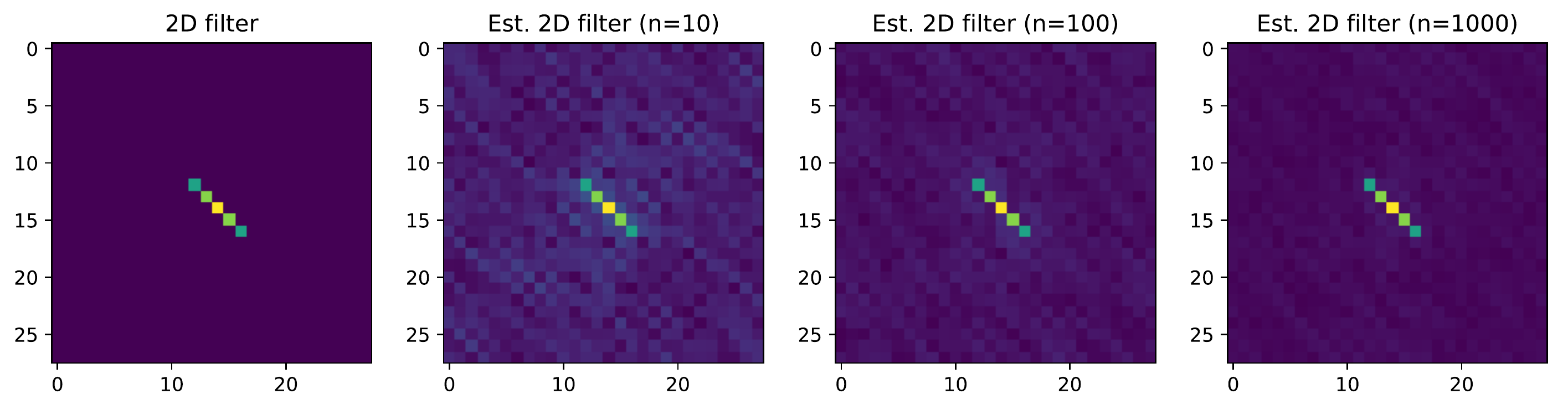}\vspace{-2mm}
    \caption{(left) 2D motion blur filter applied to the target MNIST images. (center to right)   estimated 2D filters for different number of samples $n=n_1=n_2$ for the filter estimation. All images are shown with a square root of their magnitude in order to better see small errors.}
    \label{fig:2dfilter}
\end{figure}

\begin{figure}[t]
    \centering
    \includegraphics[width=1\linewidth]{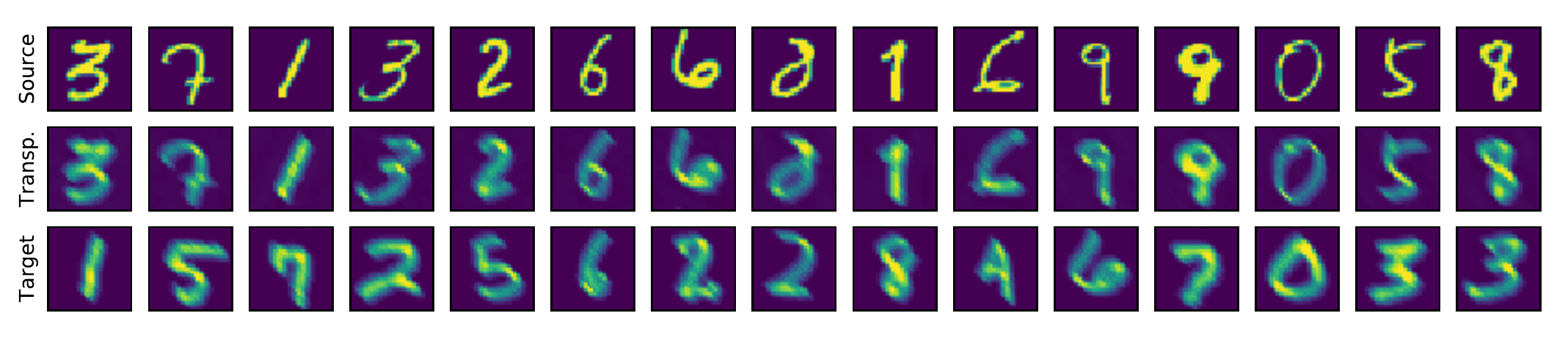}\vspace{-2mm}
    \caption{Example images from the MNIST images data. (top) Images from the source distribution. (middle) Examples from the source distribution mapped to the target distribution (convolved with estimated filter). (bottom) Examples from the Target distribution (with the motion blur filter applied).}
    \label{fig:2ddata}
\end{figure}

\paragraph{Convolutional Domain adaptation Problem.} In this section we design a
DA problem specific to image data where the transformation between the source
and target domain is a convolution operator (which is a special case of linear
operator). We design a realistic convolution operator corresponding to a motion
blur that is common on images taken in real life conditions \cite{potmesil1983modeling}. The corresponding
2D filter used in the experiment is illustrated in Figure \ref{fig:2dfilter}
(left). 
We now apply this blur filter to real images. The source/target datasets are generated as follows. We extract from the MNIST dataset two unlabeled samples sets of equal cardinality $n_1=n_2=n$. The first dataset is left unchanged and constitutes the source domain data. The blur filter of Figure \ref{fig:2dfilter} is applied to the second dataset that becomes the target domain dataset. The first line in Figure \ref{fig:2ddata} contains the source domain images and the last line the target domain ones.

\paragraph{Estimation of Monge mapping as 2D filters between distributions.} 

\begin{figure}[t]
    \centering
    \includegraphics[width=1\linewidth]{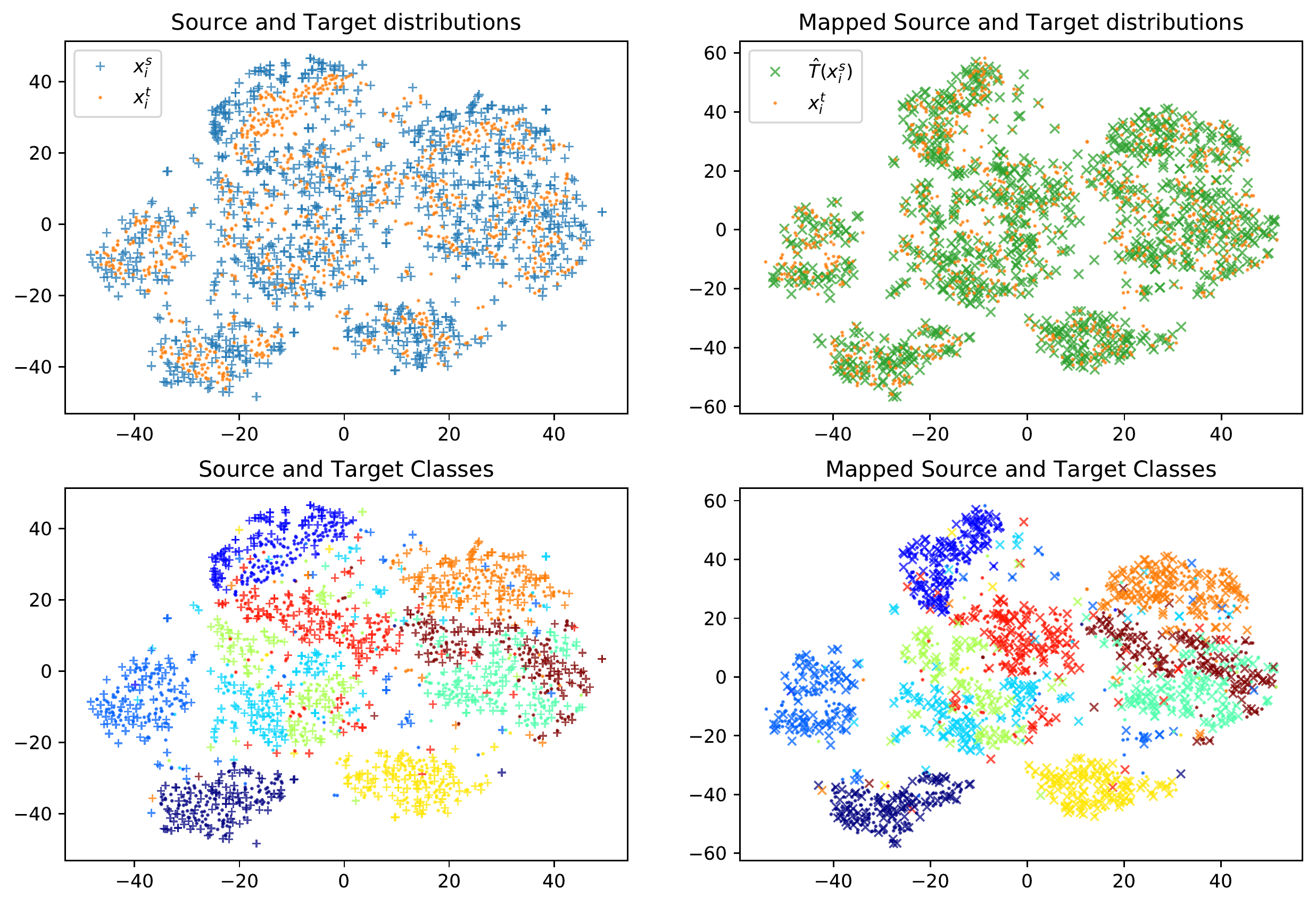}\vspace{-2mm}
    \caption{2D TSNE projection of the MNIST Source, Target and Mapped Source samples. (top line) 2D TNSE projections of Source/Target (left) and Mapped Source/Target (right) colored by distribution. (bottom line) 2D TNSE projections of Source/Target (left) and Mapped Source/Target (right) colored by class.}
    \label{fig:tsne}
\end{figure}

The 2D filters estimated using (\ref{eq:mapconv}) for a different number of
samples $n$ in source/target can
be seen in the right part of Figure \ref{fig:2dfilter}. Note that even for
$n=10$ the filter is surprisingly well estimated considering that there is not
even one sample per class in the source/target distributions. For $n=1000$ the
error on the filter is not visible anymore which is also very good for a problem
of dimensionality $d=28\times 28=784$ variables (pixels). 
Now we estimate our optimal Monge mapping using the source/target domain data and we apply this mapping to an independent sample of MNIST images in the source domain. These mapped images are represented in the
center line of Figure \ref{fig:2ddata}. We can see that
the mapped samples are very similar to the target samples. We also provide a 2D
TNSE projection \cite{maaten2008visualizing} of the samples before and after
mapping in Figure \ref{fig:tsne}. We can see that similarly to the toy data in
Figure \ref{fig:example_map} the mapping aligns the two distributions but also
the labels which suggest good domain adaptation performance.

\begin{figure}[t]
    \centering
    \includegraphics[width=.46\linewidth]{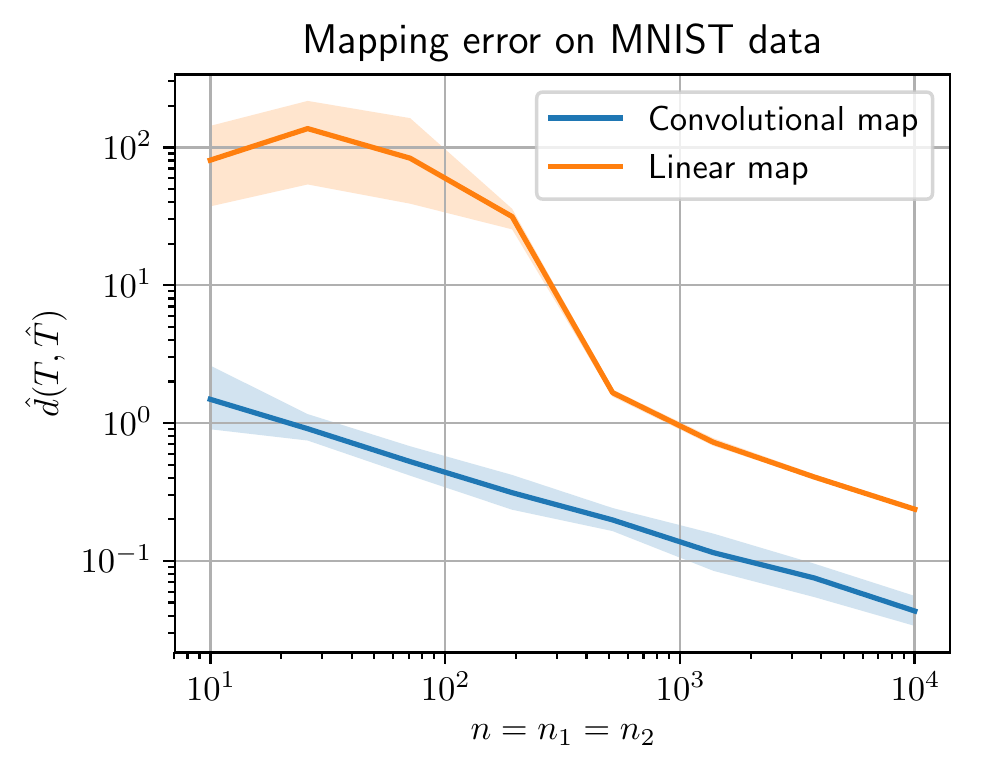}\hspace{5mm}
    \includegraphics[width=.46\linewidth]{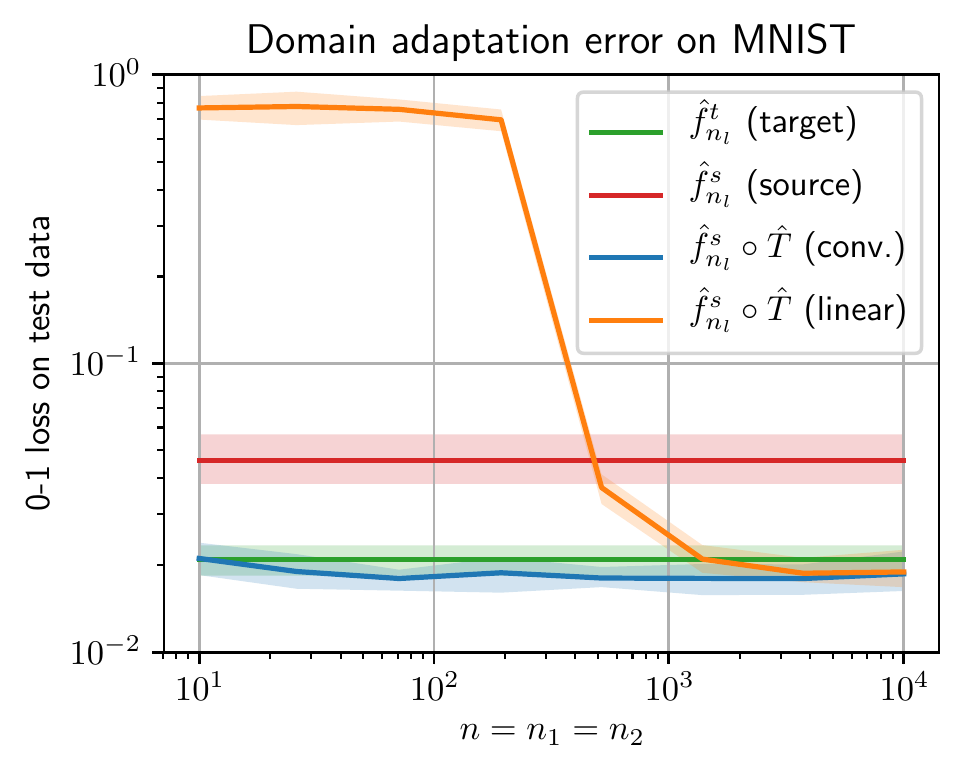}\vspace{-2mm}
    \caption{(left) Mapping error on the MNIST data for a convolutional and linear mapping (right) Target prediction error for domain adaptation problem with training on source, training on target and OTDA with linear and convolutional mapping.}
    \label{fig:err_mnist_da}
\end{figure}

In order to have a quantitative measure of the quality of both linear and
convolutional Monge mapping we perform 20 Monte Carlo experiments where we
randomly draw a varying number of $n$ samples for estimating the filter described
above. The mapping for the linear Monge mapping is estimated using the general
formulation in (\ref{eq:That}) whereas for convolutional mapping, we use the
simplified formulation in (\ref{eq:mapconv})
computed by FFT.
The average error of the mapping for both
approaches is reported in Figure \ref{fig:err_mnist_da} (left). {We can see that
the linear mapping has a hard time estimating the mapping especially when $n<d$
(estimated covariance matrix is singular but small regularization is used) but
recovers its theoretical convergence speed for $n\geq 10^{3}$. 
The convolutional mapping that is much more structured and estimates a
smaller number of parameter (block Toeplitz covariance matrix) recovers its theoretical
convergence speed for $n\geq 10$.}

\paragraph{Convolutional mapping for domain adaptation.} 

We then investigate the performance in domain adaptation  between the two domains presented
above (from original to filtered MNIST) when training a
Convolutional Neural Network (CNN) in the source domain. 
We use the architecture 
from the MNIST example of
Keras\footnote{Available at
\url{https://github.com/keras-team/keras/blob/master/examples/mnist_cnn.py}}
with all
hyperparameters fixed for all
comparisons. CNN are trained on $n_l=10^4$ labeled samples and their
test error rate is evaluated also on $10^4$ independent target samples
We first compute baselines with a CNN trained on source domain ($\hat f_{n_l}^s$) and
target domain ($\hat f_{n_l}^t$).
These baselines will allow us to assess the performance of OTDA classifier
$\hat f_{n_l}^s\circ \hat T^{-1} $ where $\hat T$ is estimated using Linear (linear) and
Convolutional (conv.) Monge Mapping estimation. 
The average classification
error on test data over 20 Monte
Carlo simulations (data sampling) with a varying $n=n_1=n_2$ is reported in Figure
\ref{fig:err_mnist_da} (right). We can see that the convolutional mapping quickly reaches
the performance of classifier $\hat f_{n_l}^t$ trained directly on
target data. {This might be due to a regularization effect coming from the
convolution operator $\hat T^{-1}$ to the data before classification that also seems to
lead to a slightly better final performance than $\hat f_{n_l}^t$.} The linear Monge
mapping requires more samples for a proper mapping ($n\geq 10^3$) estimation but also reaches
the best performance on target.

\section{Conclusion}

In this work, we provided the first concentration bound on the quality of a linear Monge
mapping when estimated from a discrete sampling. We have shown that this linear mapping
can be estimated from non-Gaussian distributions. We discussed the
computational complexity of the linear Monge mapping estimation and investigated
a variant that leads to both a speedup and better estimation when the data is a
1D/2D stationary signal which implies a convolutional mapping. This fundamental
results allowed us to prove the first bound for Optimal Transport Domain
Adaptation \citep{courty2016optimal}. This result states that OT can be used to derive a predictor in the target domain with performance converging to the Bayes risk as the sample size grows. Finally we provided numerical experiments to illustrate the theoretical bounds for both linear and convolutional mapping.

\rf{Future works will investigate the design and convergence of an applicable non-linear Monge
mapping estimated from finite distributions \citep{hutter2019minimax}.} An approach would be to study the quality of the barycentric mapping that has been used in practice
\citep{ferradans2014regularized,courty2016optimal} and is known to converge
weakly to the true Monge mapping \citep{seguy2017large}. The study of the mapping
estimation in the presence of additive noise is also an interesting research
direction related to Gaussian deconvolution \citep{rigollet2018entropic}.
Also note that the estimation of a convolutional mapping between distributions
of images opens the door for applications in image processing and especially in
astronomy where it could be used to estimate changes in the Point Spread
Function of a telescope or parameters of weak gravitational lensing~\citep{munshi2008cosmology}.

\section*{Acknowledgements}

The authors want to thank Gabriel Peyré and Nicolas Courty for fruitful discussions about linear
Monge mapping and domain adaptation. This work benefited from the support from  OATMIL ANR-17-CE23-0012 project of the French National Research Agency (ANR). 

\bibliography{refs}

\section{Proofs}
\label{sec:proof}

\subsection{Proof of Lemma \ref{thm:subgauss} }

\begin{proof}
The proof is straightforward using e.g. the lower bound on Fréchet distance between two distributions. Let consider a random vector $(x^{\top},y^{\top})^{\top}\in\mathbb{R}^{2d}$ with marginal distributions $x\sim \mu_1$ and $y\sim \mu_2$. According to \citep{Dowson1982}:
\begin{equation}
\|m_{1} - m_{2} \|^2 + \text{tr}\left(\Sigma_1 + \Sigma_2 -2 (\Sigma_1\Sigma_2)^{1/2}\right) \leq \mathbb{E} \|x-y \|^2,
\label{frechetbound}
\end{equation}
with equality if and only if $y\stackrel{d}{=}T(x) = m_2+A(x-m_1)$ where $A$ is defined as in (\ref{eq:Amap}). 
We now prove that $T$ is a legit transport map.  {First recall that the Brenier Theorem  \cite{brenier1991polar} for quadratic loss states that the optimal transport is the unique map $T$ such that $\mu_2= T_\#\mu_1$ and $T = \nabla \varphi$ for some convex function $\varphi$, see \cite[Theorem 2.32]{Villani2003}. Applying Brenier's Theorem with $\varphi(x) = (1/2)x^{\top}Ax + (m_2-A m_1)^{\top}x$ gives the result. 
}\end{proof}

\subsection{Proof of Theorem \ref{thm:ot_Monge}}
\label{sec:proof:thm1}

From now on, by abuse of notation, $\|\cdot\|$ will refer either to the $l_2$-norm of a vector or the operator norm of a matrix.

We first observe that 
\begin{align}
    \|T(x)-\hat T(x)\| & = \|\mt-\hmt +(A-\hA)( x-m_1+\hat{A}(\ms-\hms)\|\nonumber\\
     & \leq  \|\mt-\hmt\| + \|A - \hat A \|\|x - \ms\|+ \|\hA\| \|\hms - \ms \|.
\end{align}

\subsubsection{Bounding  $\|\hat{m}_j - m_j\|$, $j=1,2$}

Bounding $\|\hat{m}_j - m_j\|$, $j=1,2$ poses no particular difficulty. 
We have $X^s \stackrel{d}{=} \Sigma_1^{1/2} Z$ where $Z \in \mathbb{R}^p$ is a sub-Gaussian random vector, that is, for any deterministic vector $\alpha$, we have
$$
\mathbb{E}\left[  \exp(\langle \alpha, Z_i  \rangle) \right] \leq \exp\left(  \frac{\|\alpha\|^2}{2} \right).
$$
Note that
$$
\hat{m}_1 - m_1  \stackrel{d}{=} \Sigma_1^{1/2} \frac{1}{n_1}\sum_{i=1}^{n_1} Z_i, 
$$
where $Z_1,\ldots,Z_{n_1}$ are independent distributed as $Z$. Theorem 2.1 in \cite{hsu2012} gives for any $t>0$, with probability at least $1-e^{-t}$,
\begin{align}\label{eq:hatmj-mj}
\|\hat{m}_1 - m_1\|^2 \leq \frac{\|\Sigma_1\|}{n_1} \left[ \mathbf{r}(\Sigma_1) + 2 \sqrt{\mathbf{r}(\Sigma_1) t} + 2  t   \right]
\end{align}
A similar bound holds valid for $X^t$ with $\Sigma_1$ and $n_1$ replaced by $\Sigma_2$ and $n_2$.

\subsubsection{Bounding  $ \|A-\hA\|$ }

\paragraph{Matrix geometric mean.}
We recall first some useful facts. The geometric mean of 2 positive definite matrices is defined as
\begin{align*}
B\#C := B^{1/2}(B^{-1/2}C B^{-1/2})^{1/2} B^{1/2} = B(B^{-1}C)^{1/2}.
\end{align*} {Note that for readability, in the remaining of this section, the $\#$ operator refers to the matrix geometric mean and not to the pushforward operator $_\#$ used in the main paper.}
The matrix geometric mean satisfies
\begin{align}
    B\#C &= C\#B\label{eq:geomean1}\\
    (B\#C)^{-1} &= B^{-1} \# C^{-1}.\label{eq:geomean2}
\end{align}

We concentrate now on $\|A-\hA\|$ that requires more work. We prove the following result.
\begin{theorem}\label{thm:hA-A}
Let $\mu_1$ and $\mu_2$ be sub-Gaussian distributions with respective means and covariance $\mu_j,\Sigma_j$, $j=1,2$. Assume that
\begin{align}\label{eq:condn1}
C \left(  \sqrt{\frac{\mathbf{r}(\Sigma_1)}{n_1}} \vee \frac{\mathbf{r}(\Sigma_1)}{n_1} \vee \sqrt{\frac{\log(n_1)}{n_1}}   \right) \leq \frac{1}{2}\min\left\lbrace  \frac{1}{\kappa(\Sigma_1)},1,\frac{\lambda_{\min}(  \Sigma_1^{1/2} \Sigma_2 \Sigma_1^{1/2})}{\kappa^{1/2}(\Sigma_1)\|\Sigma_2\|}.    \right\rbrace,
\end{align}
for some sufficiently large numerical constant $C>0$.
Then we have with probability at least $1-e^{-t}-\frac{1}{n_1}$,
\begin{align}
    \|\hat A - A\| &\lesssim \frac{\kappa(\Sigma_1)\|\Sigma_2\|}{\lambda_{\min}^{1/2}(\Ss^{1/2}\St\Ss^{1/2})} \left(  \sqrt{\frac{\mathbf{r}(\Sigma_2)}{n_2}} \vee \frac{\mathbf{r}(\Sigma_2)}{n_2} \vee \sqrt{\frac{t}{n_2}} \vee \frac{t}{n_2}  \right)\nonumber\\
    &\hspace{1cm}+ \frac{\kappa(\Sigma_2)\kappa(\Sigma_1) \|\Sigma_2\| }{\lambda_{\min}^{1/2}(\Sigma_2^{-1/2}\Sigma_1\Sigma_2^{-1/2})} \left(  \sqrt{\frac{\mathbf{r}(\Sigma_1)}{n_1}} \vee \frac{\mathbf{r}(\Sigma_1)}{n_1} \vee \sqrt{\frac{\log n_1}{n_1}}   \right).
\end{align}
\end{theorem}

Based on \eqref{eq:geomean1}-\eqref{eq:geomean2}, we deduce that
\begin{align}\label{eq:hatA-A}
\hat A - A &= \hat\Sigma_1^{-1}\# \hat\Sigma_2 - \Sigma_1^{-1}\# \Sigma_2\nonumber\\
&= \hat\Sigma_1^{-1}\# \hat\Sigma_2  -  \hat\Sigma_1^{-1}\# \Sigma_2  +  \hat\Sigma_1^{-1}\# \Sigma_2   - \Sigma_1^{-1}\# \Sigma_2\nonumber\\
&=\hat\Sigma_1^{-1}\# \hat\Sigma_2  -  \hat\Sigma_1^{-1}\# \Sigma_2  +  \Sigma_2\#\hat\Sigma_1^{-1}   - \Sigma_2\#\Sigma_1^{-1}\nonumber\\
&=\hat\Sigma_1^{-1}\# \hat\Sigma_2  -  \hat\Sigma_1^{-1}\# \Sigma_2  +  (\Sigma_2^{-1}\#\hat\Sigma_1)^{-1}   - (\Sigma_2^{-1}\#\Sigma_1)^{-1}.
\end{align}

Next we have by definition of the matrix geometric mean that
\begin{align*}
    \hat\Sigma_1^{-1}\# \hat\Sigma_2  -  \hat\Sigma_1^{-1}\# \Sigma_2 
    &=\hat\Sigma_1^{-1/2} \left[ (\hat\Sigma_1^{1/2}\hat\Sigma_2 \hat\Sigma_1^{1/2})^{1/2}  -  (\hat\Sigma_1^{1/2}\Sigma_2 \hat\Sigma_1^{1/2})^{1/2}   \right] \hat\Sigma_1^{-1/2}.
    \end{align*}
Taking the operator norm, we get
    \begin{align*}\label{eq:hatA-A2}
        \| \hat\Sigma_1^{-1}\# \hat\Sigma_2  -  \hat\Sigma_1^{-1}\# \Sigma_2 \|
        &\leq \|\hat\Sigma_1^{-1/2}\|^2 \| (\hat\Sigma_1^{1/2}\hat\Sigma_2 \hat\Sigma_1^{1/2})^{1/2}  -  (\hat\Sigma_1^{1/2}\Sigma_2 \hat\Sigma_1^{1/2})^{1/2}   \|.
        \end{align*}

\paragraph{Perturbation argument.}

        We set $B = \hSs\ph\St\hSs\ph $ and $\hat{B} = \hSs\ph\hSt\hSs\ph$. 
                Note that $X = \hat{B}^{1/2} - B^{1/2}$ is solution of
        $
        XU+VX=W,
        $
        with $U = \hat{B}^{1/2}$, $V=B^{1/2}$, $W= \hat{B}-B$. Then, we can apply Lemma 2.1 in \cite{SCHMITT1992} to obtain the following bound:
        \begin{align}
        \|X\| &\leq  \frac{1}{\lambda_{\min}(B^{1/2})}\|\hat{B}-B\|.        \end{align}
        where $\lambda_{\min}(A)$ is the minimum eigenvalue of symmetric matrix $A$.

        Thus we get
\begin{align*}\label{eq:hatA-A3}
    \|X\| &\leq  \frac{1}{\lambda_{\min}(B^{1/2})}\|\hat\Sigma_1\|   \|\hat{\Sigma}_2-\Sigma_2\|.
    \end{align*}
Combining the previous display with (\ref{eq:hatA-A2}), we deduce that 
\begin{equation}\label{eq:hatA-A4}
    \| \hat\Sigma_1^{-1}\# \hat\Sigma_2  -  \hat\Sigma_1^{-1}\# \Sigma_2 \|\leq \frac{\kappa(\hat\Sigma_1)}{\lambda_{\min}((\hSs\ph\St\hSs\ph)^{1/2})} \|\hat{\Sigma}_2-\Sigma_2\|,
 \end{equation}
where $\kappa(A) =  \|A^{-1}\| \|A \|$ is the condition number of $A$.

We study now the second difference in the right-hand side of (\ref{eq:hatA-A}). In view of \cite{wedin}, we have 
$$
\|(\Sigma_2^{-1}\#\hat\Sigma_1)^{-1}   - (\Sigma_2^{-1}\#\Sigma_1)^{-1}\| \leq \|(\Sigma_2^{-1}\#\hat\Sigma_1)^{-1} \|\|(\Sigma_2^{-1}\#\Sigma_1)^{-1}\| \|  \Sigma_2^{-1}\#\hat\Sigma_1   - \Sigma_2^{-1}\#\Sigma_1 \|.
$$
A similar reasoning to that yielding (\ref{eq:hatA-A4}) gives us
\begin{align}\label{eq:hatA-A5}
\|  \Sigma_2^{-1}\#\hat\Sigma_1   - \Sigma_2^{-1}\#\Sigma_1 \|  &\leq \frac{\kappa(\Sigma_2) \|\Sigma_2\#\hat\Sigma_1^{-1} \|\|\Sigma_2\#\Sigma_1^{-1}\| }{\lambda_{\min}((\Sigma_2^{-1/2}\Sigma_1 \Sigma_2^{-1/2})^{1/2})} \| \hat \Sigma_1 - \Sigma_1 \|.
\end{align}
Combining the last display with (\ref{eq:hatA-A}) and (\ref{eq:hatA-A4}), we obtain 
\begin{align}\label{eq:hatA-A5}
\|\hat A - A\|  &\leq \frac{\kappa(\hat\Sigma_1)}{\lambda_{\min}^{1/2}(\hSs\ph\St\hSs\ph)} \|E_2\|+ \frac{\kappa(\Sigma_2) \|\Sigma_2\#\hat\Sigma_1^{-1} \|\|\Sigma_2\#\Sigma_1^{-1}\| }{\lambda_{\min}^{1/2}(\Sigma_2^{-1/2}\Sigma_1 \Sigma_2^{-1/2})} \| E_1 \|.
    \end{align}
with $E_1 := \hat \Sigma_1 - \Sigma_1$ and $E_2 := \hat{\Sigma}_2-\Sigma_2$.

We now need to control the following randon terms: $\kappa(\hat\Sigma_1)$, $\lambda_{\min}((\hSs\ph\St\hSs\ph)^{1/2})$ and $\|\Sigma_2\#\hat\Sigma_1^{-1} \|$. To this end, we introduce the event
\begin{align}
\mathcal E_1 = \left\{ \|\Sigma_1^{-1} E_1 \| \leq   \frac{1}{2}   \right\} \cap \left\{ \|E_1 \| \leq   \frac{\|\Sigma_1\|}{2}   \right\} \cap \left\{   \|E_1\| \leq \frac{\lambda_{\min}(\Sigma_1^{1/2}\Sigma_2 \Sigma_1^{1/2})}{2 \sqrt{6}\|\Sigma_2\| \kappa^{1/2}(\Sigma_1)}    \right\}.
\end{align}
We have on $\mathcal E_1 $ that
$$
\| \hat \Sigma_1^{-1} - \Sigma_1^{-1} \| \leq 2 \| \Sigma_1^{-1} E_1\| \|\Sigma_1^{-1}\|,
$$
and consequently
$$
\|\hat\Sigma_1^{-1}\| \leq 2 \| \Sigma_1^{-1} \|,\quad \|\hat\Sigma_1 \| \leq \frac{3}{2} \|\Sigma_1\|.
 $$
Thus we have on $\mathcal E_1$ that 
$$\kappa(\hat \Sigma_1) \leq 3 \kappa(\Sigma_1)$$
and 
$$
\|\Sigma_2 \# \hat\Sigma_1^{-1}\| \leq \|\Sigma_2\|^{1/2} \|\hat \Sigma_1^{-1}\|^{1/2} \leq \sqrt{2}\|\Sigma_2\|^{1/2}\|\Sigma_1^{-1}\|^{1/2}.
$$
Applying again Lemma 2.1 in \cite{SCHMITT1992}, we get that 
\begin{align*}
&\left| \lambda_{\min}((\hat\Sigma_1^{1/2}\Sigma_2 \hat\Sigma_1^{1/2})^{1/2})  - \lambda_{\min}((\Sigma_1^{1/2}\Sigma_2 \Sigma_1^{1/2})^{1/2}) \right| \\
&\hspace{5cm}\leq \frac{1}{\lambda_{\min}^{1/2}(\Sigma_1^{1/2}\Sigma_2\Sigma_1^{1/2})}\|   \hat\Sigma_1^{1/2}\Sigma_2 \hat\Sigma_1^{1/2}   -  \Sigma_1^{1/2}\Sigma_2\Sigma_1^{1/2}   \|.
\end{align*}
Next we note that
$$
\| \hat\Sigma_1^{1/2}\Sigma_2 \hat\Sigma_1^{1/2}   -  \Sigma_1^{1/2}\Sigma_2\Sigma_1^{1/2}   \| \leq 2 \|\hat\Sigma_1^{1/2}\Sigma_2\| \|\hat\Sigma_1^{1/2} - \Sigma_1^{1/2}\|.
$$
We apply again Lemma 2.1 in \cite{SCHMITT1992} to get
$$
\|\hat\Sigma_1^{1/2} - \Sigma_1^{1/2}\| \leq \frac{1}{\lambda_{\min}^{1/2}(\Sigma_1)} \|  \hat\Sigma_1 - \Sigma_1 \|.
$$
Combining the last three displays, we get on the event $\mathcal E_1$ that
\begin{align*}
&\left| \lambda_{\min}((\hat\Sigma_1^{1/2}\Sigma_2 \hat\Sigma_1^{1/2})^{1/2})  - \lambda_{\min}((\Sigma_1^{1/2}\Sigma_2 \Sigma_1^{1/2})^{1/2}) \right| \\
&\hspace{5cm}\leq \sqrt{6}\frac{\|\Sigma_1\|^{1/2}\|\Sigma_2\|}{\lambda_{\min}^{1/2}(\Sigma_1^{1/2}\Sigma_2\Sigma_1^{1/2})   \lambda_{\min}^{1/2}(\Sigma_1)}\|  E_1   \|\\
&\hspace{5cm}\leq \frac{1}{2} \lambda_{\min}((\Sigma_1^{1/2}\Sigma_2 \Sigma_1^{1/2})^{1/2}).
\end{align*}

Thus, we get on the event $\mathcal E_1$ that
$$
\lambda_{\min}^{1/2}(\hat\Sigma_1^{1/2}\Sigma_2 \hat\Sigma_1^{1/2}) \geq \frac{1}{2}\lambda_{\min}^{1/2}(\Sigma_1^{1/2}\Sigma_2\Sigma_1^{1/2} ).
$$

Combining these facts with (\ref{eq:hatA-A5}), we get
\begin{align}\label{eq:hatA-A6}
    \|\hat A - A\|  &\lesssim \frac{\kappa(\Sigma_1)}{\lambda_{\min}^{1/2}(\Sigma_1^{1/2}\St\Sigma_1^{1/2})} \|E_2\|+ \frac{\kappa(\Sigma_2) \|\Sigma_2\|\|\Sigma_1^{-1} \| }{\lambda_{\min}^{1/2}(\Sigma_2^{-1/2}\Sigma_1\Sigma_2^{-1/2})} \| E_1 \|.
\end{align}

\paragraph{Bounding $\left\| E_1 \right\|$ and $\left\| E_2 \right\|$.}
We treat $\|E_1\|$. The result for $\left\| E_2 \right\|$ follows from the same argument. We set $Y_i = X_i - m_1$, $1\leq i \leq n$. We have
\begin{align*}
\hat \Sigma_1 -  \Sigma_1 &= \frac{1}{n_1}\sum_{i=1}^{n_1} (X_i - \bar{X})(X_i - \hat{m}_1)^\top\\
&= \frac{1}{n_1} \sum_{i=1}^{n_1} Y_i Y_i^\top - \Sigma_1   -2 (m_1 - \hat{m_1}) (m_1 - \hat{m_1})^\top. 
\end{align*}
We get from the previous display
\begin{align*}
\|\hat \Sigma_1 -  \Sigma_1\| &\leq \left\| \frac{1}{n_1} \sum_{i=1}^{n_1} Y_i Y_i^\top - \Sigma_1\right\| +2 \left\|  (m_1 - \hat{m_1}) (m_1 - \hat{m_1})^\top\right\|\\
&\leq \left\| \frac{1}{n_1} \sum_{i=1}^{n_1} Y_i Y_i^\top - \Sigma_1\right\| +2\left\|   m_1 - \hat{m_1}  \right\|^2.
\end{align*}
The second term in the right hand side of the previous display was already treated in \eqref{eq:hatmj-mj}. We apply now Theorem 2 in \cite{KLIHP} to handle the first term. We obtain for any $t>0$, with probability at least $1-e^{-t}$,
$$
 \left\| \frac{1}{n_1} \sum_{i=1}^{n_1} Y_i Y_i^\top - \Sigma_1\right\| \lesssim \|\Sigma_1\| \left(\sqrt{\frac{\mathbf{r}(\Sigma_1)}{n_1}}\bigvee \frac{\mathbf{r}(\Sigma_1)}{n_1} \bigvee \sqrt{\frac{t}{n_1}} \bigvee \frac{t}{n_1}\right).
$$
Combining the previous display with \eqref{eq:hatmj-mj}, we  get (up to a rescaling of the constants) for any $t>0$, with probability at least $1-e^{-t}$,
$$
 \left\| E_1 \right\| \leq C \|\Sigma_1\| \left(\sqrt{\frac{\mathbf{r}(\Sigma_1)}{n_1}}\bigvee \frac{\mathbf{r}(\Sigma_1)}{n_1} \bigvee \sqrt{\frac{t}{n_1}} \bigvee \frac{t}{n_1}\right).
$$
for some sufficiently large numerical constant $C>0$.
Taking $t_1 = 2 \log n_1$ and using condition (\ref{eq:condn1}), we obtain the result.

\medskip

Now lets go back to the original problem. We have
\begin{align}
    \|T(x)-\hat T(x)\| & \leq  \|\mt-\hmt\| + \|A - \hat A \|\|x - \ms\|+ \|\hA\| \|\hms - \ms \|
\end{align}
Taking the expectation w.r.t. $x\sim \mu_1$, we get
\begin{align}
    \E_{x\sim \mu_1} \left[\|T(x)-\hat T(x)\| \right] & \leq  \|\mt-\hmt\|  +  \|A - \hat A \|\mathbb{E}_{x\sim \mu_1}\left[   \|x - \ms\|\right]+ \|\hA\| \|\hms - \ms \| \nonumber\\
& \leq  \|\mt-\hmt\|  +  \|A - \hat A \|\mathbb{E}_{x\sim \mu_1}^{1/2}\left[   \|x - \ms\|^2\right]+ \|\hA\| \|\hms - \ms \|, \nonumber\\
\end{align}
where we use Cauchy-Schwarz's inequality in the last line.

Bounding $\|\mathbb{E}_{x\sim \mu_1}^{1/2}\left[   \|x - \ms\|^2\right]$ is a straightforward computation. We get
$$
\mathbb{E}_{x\sim \mu_1}^{1/2}\left[   \|x - \ms\|^2\right]  \leq C\|\Sigma_1\|^{1/2} \sqrt{\mathbf{r}(\Sigma_1)},
$$
for some numerical constant $C>0$.

Under conditions \supp{(\ref{cond-1-Sigma_j}), (\ref{cond-2-Sigma_j})}{(5), (6)} with Theorem \ref{thm:hA-A}, we get that 
$$
\|\hat A\|  \leq_{\mathbb{P}} \|A\| + \|\hat A - A\| \leq C
$$
for some numerical constant $C>0$.

Combining (\ref{eq:hatmj-mj}), Theorem \ref{thm:hA-A} with conditions \supp{(\ref{cond-1-Sigma_j}), (\ref{cond-2-Sigma_j})}{(5), (6)}, we get with
probability at least $1-3 e^{-t}-\frac{1}{n_1}$,
\begin{align}
    d(T,\hat{T}) \leq  C'   \left(      \sqrt{\frac{\mathbf{r}(\Sigma_2)}{n_2}} \vee \sqrt{\frac{\mathbf{r}(\Sigma_1)}{n_1}} \vee   \sqrt{\frac{t}{n_1 \wedge n_2}} \vee \frac{t}{n_1\wedge n_2} \right) \sqrt{\mathbf{r}(\Sigma_1)} ,
\end{align}
for some absolute constant $C'>0$. Up to a rescaling of the constant, we can replace probability $1-3 e^{-t}-\frac{1}{n_1}$ by $1- e^{-t}-\frac{1}{n_1}$.

\subsection{Proof of Proposition \ref{thm:dapred}}
\label{sec:proof_2}

In view of \supp{(\ref{eq:assumptionotda})}{(11)}, we have
\begin{align}
&\mathbb{E}_{(x,y)\sim \mathcal{P}_t} \left[   L(y, f\circ \hat T^{-1}(x))    \right]\nonumber\\
&\hspace{1.5cm}=  \mathbb{E}_{(x,y)\sim \mathcal{P}_s} \left[   L(y, f\circ \hat T^{-1}(T(x)))    \right]\nonumber\\
&\hspace{1.5cm}= \mathbb{E}_{(x,y)\sim \mathcal{P}_s} \left[   L(y, f\circ \hat T^{-1}(\hat T(x)))    \right]+ \mathbb{E}_{(x,y)\sim \mathcal{P}_s} \left[   L(y, f\circ \hat T^{-1}(T(x)))  -  L(y, f\circ \hat T^{-1}(\hat T(x)))    \right]\nonumber\\
&\hspace{1.5cm}\leq \mathbb{E}_{(x,y)\sim \mathcal{P}_s} \left[   L(y, f (x))    \right] + M_f M_L \mathbb{E}_{(x,y)\sim \mathcal{P}_s} \left[   \| \hat T^{-1}(T(x)) -  \hat T^{-1}(\hat T(x)) \|   \right]\label{DA-oracleinequality}\\
&\hspace{1.5cm}\leq \mathbb{E}_{(x,y)\sim \mathcal{P}_s} \left[   L(y, f (x))    \right] + M_f M_L \mathbb{E}_{(x,y)\sim \mathcal{P}_s} \left[   \| \widehat{A}^{-1}(T(x) -  \hat T(x)) \|   \right]\nonumber\\
&\hspace{1.5cm}\leq R_s(f) + M_f M_L  \mathbb{E}_{(x,y)\sim \mathcal{P}_s} \left[  \|\widehat{A}^{-1}\|\| T(x) -  \hat T(x) \|   \right]\notag\\
&\hspace{1.5cm}\leq R_s(f) + M_f M_L  \|\widehat{A}^{-1}\|\, d(T,\hat{T}),\label{DA-oracleinequalityMP}
\end{align} where the last two lines follows from the definition of $\hat T^{-1}$.

\subsection{Proof of theorem 3}
\label{sec:proof_3}

Let $\mathcal{H}_K$ be a reproducing kernel Hilbert space (RKHS) associated with a symmetric nonnegatively definite kernel $K\,:\, \mathbb{R}^d \times \mathbb{R}^d \rightarrow \mathbb{R}$ such that
for any $x\in \mathbb{R}^d$, $K_x(\cdot) = K(\cdot,x) \in \mathcal{H}_K$ and $f(x) = \langle f(x),K_x \rangle_{\mathcal{H}_K}$ {for all $f \in \mathcal{H}_K$}. See the seminal paper \cite{Aronszajn} for more details. Let $\Pi_s$ be the marginal distribution of $X^s$ and $T_K$ be the integral operator from $L_2(\Pi_s)$ into $L_2(\Pi_s)$ with square integrable kernel $K$.
Then it is known that the operator $T_K$ is compact, self-adjoint and its spectrum is discrete. Let $\{\lambda_k\}_{k\geq 1}$ be the eigenvalues of $T_K$ arranged in decreasing order and $\{ \phi_k \}$ are the corresponding $L_2(\Pi_s)$-orthonormal eigenfunctions.
Then the RKHS-norm of any function $f$ in the linear span of $\{\phi_k\}$ can be written as
$$
\| f  \|_{\mathcal H_K}^2 = \sum_{k\geq 1} \frac{|\langle  f , \phi_k \rangle_{L_2(\Pi_s)}|^2}{\lambda_k}.
$$
Set $f_*^s = \mathrm{argmin}_{f} R_s(f)$. Assume that $f_* \in \mathcal H_K$ and $\|f_*\|_{\mathcal H_K}\leq 1$. We consider the following empirical risk minimization estimator:
\begin{align}
    \hat f_{n_l} := \mathrm{argmin}_{\|f\|_{\mathcal{H}_K}\leq 1} \frac{1}{n_l}\sum_{i=1}^{n_l} l(Y^l_i, f(X^l_i)).
\end{align}
The performances of this procedure have been investigated in \cite{mendelson}. If we assume in particular that $\lambda_k \asymp k^{-2\beta}$ for some $\beta>1/2$, then there exists a constant 
$C>0$ such that with probability at least $1-e^{-t}$,
$$
R_s(\hat f_n) \leq R_s(f_*^s) + C \left( n_l^{-2\beta/(1+2\beta)}  + \frac{t}{n_l}  \right)
$$

Under the assumptions of Theorem \supp{\ref{thm:ot_Monge}}{1}, we have $\|\widehat{A}^{-1}\| \leq_{\mathbb{P}} C$ 
for some numerical constant $C>0$. Combining the previous display with (\ref{DA-oracleinequalityMP}) 
and Theorem \supp{\ref{thm:ot_Monge}}{1}, we get with probability at least $1-e^{-t}-\frac{1}{n_1}$,
\begin{align}
R_t(\hat f_{n_l} \circ \hat T^{-1}) - R_s(f_*^s) &\lesssim   n_l^{-2\beta/(1+2\beta)}  + \frac{t}{n_l} \nonumber\\
&\hspace{1cm} +  M_L  \left(      \sqrt{\frac{\mathbf{r}(\Sigma_2)}{n_2}} \vee \sqrt{\frac{\mathbf{r}(\Sigma_1)}{n_1}} \vee   \sqrt{\frac{t}{n_1 \wedge n_2}} \vee \frac{t}{n_1\wedge n_2} \right) \sqrt{\mathbf{r}(\Sigma_1)} .
\end{align}
The final bound use the fact that with our assumption $R_s(f_*^s)=R_t(f_*^t)$.

\bigskip

\end{document}